\documentclass{article}
\usepackage{graphicx}
\usepackage{spconf}
\usepackage{amsfonts}
\usepackage{amsmath}
\usepackage{amssymb}
\usepackage{float}
\usepackage[justification=raggedright, singlelinecheck=false]{caption}
\usepackage{algorithm}
\usepackage{algorithmic}
\usepackage{xcolor}
\usepackage{epstopdf}
\usepackage{array}
\usepackage{comment}
\usepackage{multirow}
\usepackage{tabularx}
\usepackage{makecell}
\usepackage{soul}
\usepackage{enumitem}

\title{A text-to-text Alignment Algorithm for\\better evaluation of modern speech recognition systems}
\name{Lasse Borgholt,\textsuperscript{1,2,3} Jakob Havtorn,\textsuperscript{1} Christian Igel,\textsuperscript{3,5} Lars Maaløe,\textsuperscript{1,4} and Zheng-Hua Tan\textsuperscript{ 2,3}
    \thanks{Supported by Innovation Fund Denmark (grant no. 2051-00015B)}}
\address{\textsuperscript{1}Corti, Copenhagen \\
        \textsuperscript{2}Aalborg University, Department of Electronic Systems \\
        \textsuperscript{3}Pioneer Centre for Artificial Intelligence, Copenhagen \\
        \textsuperscript{4}Technical University of Denmark, Department of Applied Mathematics and Computer Science \\
        \textsuperscript{5}University of Copenhagen, Department of Computer Science \\
        lb@corti.ai}
\date{August 2025}

\begin{document}

%
\maketitle
\begin{abstract}
Modern neural networks have greatly improved performance across speech recognition benchmarks. However, gains are often driven by frequent words with limited semantic weight, which can obscure meaningful differences in word error rate, the primary evaluation metric. Errors in rare terms, named entities, and domain-specific vocabulary are more consequential, but remain hidden by aggregate metrics. This highlights the need for finer-grained error analysis, which depends on accurate alignment between reference and model transcripts. However, conventional alignment methods are not designed for such precision. We propose a novel alignment algorithm that couples dynamic programming with beam search scoring. Compared to traditional text alignment methods, our approach provides more accurate alignment of individual errors, enabling reliable error analysis. The algorithm is made available via PyPI.\footnote{https://github.com/corticph/error-align}

\end{abstract}
\vspace{-0.2em}
\begin{keywords}
Automatic speech recognition, text alignment, error analysis, dynamic programming, beam search
\end{keywords}
\section{Introduction}
\label{sec:intro}

\newcommand{\sub}[1]{\sethlcolor{yellow!12}\hl{#1}} 
\newcommand{\ins}[1]{\sethlcolor{blue!8}\hl{#1}} 
\newcommand{\del}[1]{\sethlcolor{red!8}\hl{#1}}    

\newcommand{\subword}{\sethlcolor{yellow!12}\hl{\textsc{sub}}} 
\newcommand{\insword}{\sethlcolor{blue!8}\hl{\textsc{ins}}}    
\newcommand{\delword}{\sethlcolor{red!8}\hl{\textsc{del}}}     
\newcommand{\matchword}{\sethlcolor{green!8}\hl{\textsc{match}}}     

\begin{table*}[h]
\centering
\renewcommand{\arraystretch}{1.2} 
\setlength{\tabcolsep}{6pt}       

\newcolumntype{Y}{>{\centering\arraybackslash}X}
\newcommand{\rc}[2]{\makecell{\scriptsize{#1 (#2\%)}}}

\begin{tabularx}{\textwidth}{|c|Y|Y|}

\hline
& {\small\textit{Levenshtein-based word-level alignment} {\scriptsize(1 of 11 equal paths)}} & {\small\textit{Our approach} {\scriptsize(1 best path)}} \\
\hline
{\small\textsc{ref}} & {\small\texttt{|some|\phantom{x}things\phantom{xx}|\phantom{x}are\phantom{x}|\phantom{x}worth\phantom{x}|noting|}} & {\small\texttt{|some\phantom{x}|things|are|worth|noting\phantom{x}|\phantom{xxxxxx}|}} \\

\noalign{\vspace{-0.4em}}
& {\small\texttt{\phantom{x}\delword\phantom{xxxxx}\subword\phantom{xxxxx}\subword\phantom{xxxx}\subword\phantom{xxxx}\subword\phantom{xxx}}} & {\small\texttt{\phantom{xx}\subword\phantom{xxx}\subword\phantom{xxx}\delword\phantom{x}\matchword\phantom{xxx}\subword\phantom{xxxx}\insword\phantom{xxx}}} \\
\noalign{\vspace{-0.4em}}

{\small\textsc{hyp}} & {\small\texttt{|\phantom{xxxx}|something|worth|nothing|period|}} & {\small\texttt{|some-|-thing|\phantom{xxx}|worth|nothing|period|}} \\

\hline
\end{tabularx}
\vspace{-0.4em}
\caption{Handcrafted example: While both alignments result in five errors, our approach capture more plausible alignments.}
\vspace{-0.8em}
\label{tab:alignment_examples}
\end{table*}

Recent advances in neural architectures and large-scale weakly-supervised training have enabled automatic speech recognition (ASR) systems to reach unprecedented accuracy \cite{radford2023robust, puvvada2024less, saon2025granite}. However, evaluation methods have not kept pace. The \textit{word error rate} (WER), based on Levenshtein distance, remains the de facto standard for benchmarking performance. While WER provides a simple and interpretable summary metric, there is a need for more fine-grained error analysis to better understand, diagnose, and improve model behavior.

Reliable error analysis requires accurate alignment between words in the reference transcript and the model transcript. High-quality alignments allow researchers and practitioners to query the model output for specific words or phrases that were mistranscribed, enabling rapid assessment of error severity for critical vocabulary items. Levenshtein-based alignment is widely used for this purpose, but as we will show, it often fails to capture plausible correspondences.

We address this limitation by proposing a new alignment approach that integrates dynamic programming with beam search scoring. The contributions of this work are threefold:
\vspace{-0.2em}
\begin{itemize}[itemsep=-0.1em]
\item We identify key limitations of Levenshtein-based alignment when applied to speech recognition outputs.
\item We propose a robust two-pass alignment algorithm that leverages character-level features and structured transition costs to yield more interpretable alignments.
\item We empirically evaluate the method across multiple models and languages, showing consistent improvements over conventional techniques and previous work.
\end{itemize}

\vspace{-0.2em}
\section{Background and Challenges}
\label{sec:challenges}


In speech recognition, alignment refers to mapping morphological units in a gold-standard manual transcript (\textit{reference}) to those in a model-generated transcript (\textit{hypothesis}).

Consider a single reference–hypothesis pair $(r, h)$. We define a valid alignment $a$ between the two strings as a sequence of index range pairs, such that $a_n = (i\!:\!j, k\!:\!l)$. Following the edit distance convention, each mapping may represent an insertion ($r_{i:j} = \varepsilon$), deletion ($h_{k:l} = \varepsilon$), substitution ($h_{k:l} \neq r_{i:j}$), or match ($h_{k:l} = r_{i:j}$). Index ranges must increase monotonically between consecutive pairs $a_n$ and $a_{n+1}$ for both the reference and the hypothesis. Finally, the alignment must cover all voiced characters, while unvoiced characters such as spaces and punctuation can be ignored.

For our purposes, each reference substring $r_{i:j}$ must correspond to exactly one word. This restriction enables users to query instances of individual words or predefined vocabularies, facilitating fine-grained error analysis. The proposed algorithm can be configured to capture phrases, if desirable.

\vspace{-0.2em}
\subsection{Levenshtein-based alignment}
\label{sec:levenshtein}

The Levenshtein distance \cite{levenshtein1966binary} is computed using a dynamic programming table, where the optimal value is determined recursively by comparing all positions in the two input sequences \cite{wagner1974string}. The corresponding alignment can be extracted through a \textit{backtrace} constructed by recording the path taken to reach each cell in the table. However, the minimum distance might be attained by multiple distinct paths, which introduces ambiguity into the resulting alignment.

\vspace{-0.2em}
\subsection{Failure cases of word-level Levenshtein}
\label{ssec:word_level}

We identify two primary failure cases of Levenshtein-based word-level alignment. First, substitutions are constrained to be strictly one-to-one. In other words, a reference word cannot be mapped to multiple words or subwords in the hypothesis. This limitation is especially problematic for agglutinative languages or languages with a high degree of noun compounding. For example, in Table~\ref{tab:alignment_examples}, the phrase \texttt{some things} is incorrectly transcribed as \texttt{something}, but is incorrectly aligned by the word-level method.

Second, substitutions that occur adjacent to insertions or deletions introduce ambiguity. Returning to the example in Table~\ref{tab:alignment_examples}, the Levenshtein-based alignment produces four substitutions and a single deletion. However, any of the reference words could have been labeled as the deletion, yielding five distinct alignments with the same optimal Levenshtein distance. This demonstrates that treating words as canonical units is insufficient when aiming for unambiguous alignment.

\vspace{-0.2em}
\subsection{Failure cases of character-level Levenshtein}
\label{ssec:character_level}

From the examples above, the lack of character-level information is clearly a key limitation of word-level alignment.
A better approach is to work fully at the character level, though plain character-level alignment also has drawbacks.
First, our goal is to map each reference word to a corresponding segment of the hypothesis transcript — something that character-level alignment does not provide.
Second, this approach offers no measure of proximity between matched characters within a single word. For example, if a model fails to capture the last letter of a word and proceeds to hallucinate a sequence of words containing the missing letter, it will be erroneously matched to the hallucinated part, resulting in a highly implausible alignment that may span multiple words.

\vspace{-0.2em}

\begin{figure}[b!]
    \centering
    \includegraphics[width=0.98\columnwidth]{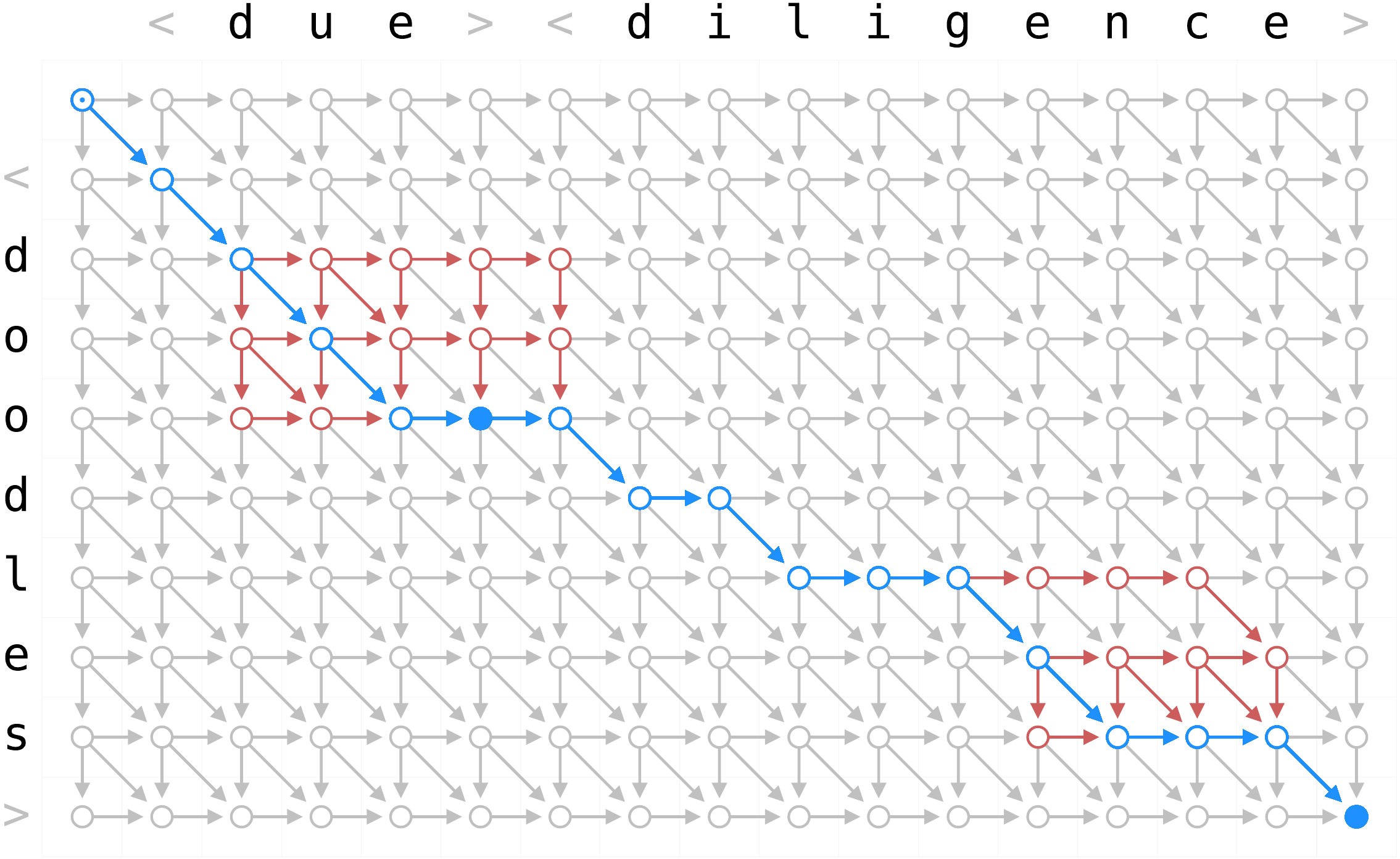}
    \vspace{-0.8em}
    \caption{Our algorithm searches the full graph (gray), penalizing deviations from the backtrace $G_b$ (red), to find the best path (blue). Solid nodes mark segment ends, eq. \eqref{eq:end_segment}.}
    \label{fig:alignment_graph}
\end{figure}

\section{Method}
\label{sec:method}
\vspace{-0.2em}

The table commonly used for alignment via dynamic programming can also be viewed as a directed acyclic graph, where cells correspond to nodes and insertions, deletions, and substitutions or matches are represented by vertical, horizontal, and diagonal edges, respectively. The backtrace defines a subgraph that captures all optimal alignments. See Figure~\ref{fig:alignment_graph}.

Our objective is to identify the minimal-cost path through this graph and establish a ruleset for mapping character-level operations to word-level alignments. The proposed algorithm leverages the backtrace subgraph as an anchor for a beam search over the full graph.

\vspace{-0.3em}
\subsection{Text pre-processing}
\label{ssec:preprocessing}
\vspace{-0.2em}

Word tokens are normalized by lower casing, removing diacritics, and replacing unvoiced characters, such as apostrophes and hyphens, with a placeholder symbol (\texttt{\#}). Finally, words are enclosed in angle brackets (i.e., \texttt{<word>}), which we use for our ruleset for extracting word-level alignments from character-level input.

\vspace{-0.3em}
\subsection{Backtrace graph extraction}
\label{backtrace}
\vspace{-0.2em}

We find that an unrestricted search over the entire graph is computationally infeasible. To prioritize the most promising paths, we introduce a penalty term when a path deviates from the backtrace graph $G_b$, which substantially improves both the robustness and efficiency of the beam search.

For graph construction, we adopt a slightly modified version of the standard unit-cost Levenshtein distance by doubling the cost of substitutions. In this formulation, every substitution can equivalently be represented as a deletion–insertion pair, which has the  effect of expanding the backtrace graph. Relaxing the unit-cost constraint on the search space in this way leads to more accurate alignments.

\renewcommand{\arraystretch}{1.1}

\newcolumntype{Y}{>{\arraybackslash}X} 
\newcolumntype{Z}{>{\centering\arraybackslash}m{0.8cm}} 

\begin{table}[t!]
\centering
\begin{tabularx}{\columnwidth}{|Z|Y|}
\hline
\textit{Cost} & \textit{Transition description} \\ \hline
0 & Match of any characters. \\ \hline
1 & Insertion or deletion of unvoiced characters.\\ \hline
2 & Insertion or deletion of voiced character. \\ \hline
2 & Substitution of vowel or consonant pair. \\ \hline
3 & Substitution of vowel and consonant. \\ \hline
\end{tabularx}
\caption{Description of the transition rules from eq. \eqref{eq:transition}. Substitutions involving unvoiced characters are prohibited.}
\label{tab:nl_costs}
\end{table}

\vspace{-0.4em}
\subsection{Path scoring}
\label{sssec:beamsearch_scoring}

We initialize the beam with a single candidate path $p$ starting at the root node $(0,0)$. At each iteration, candidates expand by transitioning to their child nodes. Word-level alignments are traced out as the candidate paths traverse the graph.

If we consider a path at node $v = (i, j)$ with the last word-level alignment of that path ending at node $u = (k, l)$, the normalized cost is defined as
\begin{align}
c = \frac{c_c + c_o(1 + [(i-k)(j-l) > 0])}{i + j + 1} \quad, \label{eq:cost}
\end{align}
\noindent where $c_c$ denotes the cumulative cost of \textit{closed} word-level alignments, $c_o$ is the cost of the current \textit{open} word-level alignment, and $i+j+1$ is a normalization term accounting for the number of characters covered thus far. The indicator function $[(i-k)(j-l) > 0]$ imposes a penalty by doubling $c_o$ for substitutions, ensuring that dissimilar words are not aligned. Importantly, because this penalty doubles the cost of previous actions accumulated by $c_o$, the problem does not have an optimal substructure, which is why a strict dynamic programming solution is not possible.

When a segment is closed, $c_c$ is updated and $c_o$ is reset:
\begin{align}
c_c &\leftarrow c_c + c_o(1 + [(i-k)(j-l) > 0]) \label{eq:close_1} \quad, \\
c_o &\leftarrow 0 \label{eq:close_2} \quad.
\end{align}
The open cost $c_o$ is updated at each iteration when transitioning to a new child node:
\begin{align}
c_o \leftarrow c_o + t_{w \rightarrow v} + [w \notin V(G_b)] \quad, \label{eq:open_update}
\end{align}

\noindent where $t_{w \rightarrow v}$ is the cost of transitioning from node $w=(m,n)$ to $v=(i,j)$ and $[w \notin V(G_b)]$ is an indicator function penalizing deviations from the backtrace graph node set.
 
For the transition cost, we use a coarse phonemic classification $P(\cdot)$ by grouping characters into vowels and consonants. In addition, the set of unvoiced characters $U = \{\texttt{<}, \texttt{>}, \texttt{\#}\}$ is subject to separate rules:
\begin{align}
t_{w \rightarrow v} =
\begin{cases}
0, & \text{if } \searrow \land \; r_j = h_i, \\
1, & \text{if } (\rightarrow \land \; h_i \in U) \lor (\downarrow \land \; r_j \in U),  \\
2, & \text{if } (\rightarrow \land \; h_i \notin U) \lor (\downarrow \land \; r_j \notin U),  \\
2, & \text{if } \searrow \land \; P(r_j) = P(h_i), \\
3, & \text{if } \searrow \land \; P(r_j) \neq P(h_i). \\
\end{cases} \label{eq:transition}
\end{align}

\noindent The rules are summarized in natural language in Table \ref{tab:nl_costs}. The arrow notation is a symbolic shorthand for graph transitions and is logically equivalent to
\begin{align}
\rightarrow&\;\Leftrightarrow v-w=(0, 1) \quad, \\
\downarrow&\;\Leftrightarrow v-w=(1, 0) \quad, \\
\searrow&\;\Leftrightarrow v-w=(1, 1) \quad.
\end{align}


To determine if a segment should be closed, we define a function that is evaluated at each transition and updates $u$, if the alignment is to be concluded. Again, consider a path transitioning from $w$ to $v$ with $u$ as the end node of the previously recorded alignment. We then have
\begin{align}
u \leftarrow
\begin{cases}
v, & \text{if } (\searrow \lor \rightarrow) \land r_j = \texttt{>}, \\
w, & \text{if } (\searrow \lor \rightarrow)  \land r_j = \texttt{<} \land w \neq u, \\
v, & \text{if } \downarrow \land \; h_i = \texttt{>} \land j=l \land w \neq u, \\
u, & \text{otherwise,} \label{eq:end_segment}
\end{cases}
\end{align}

\noindent where \texttt{<} and \texttt{>} are the start- and end-of-token delimiters. If the output is not $u$, the segment is closed, the word-level alignment is recorded, and equations \eqref{eq:close_1} and \eqref{eq:close_2} are applied.

\begin{table*}[htbp]
\begin{minipage}[t]{0.60\linewidth}
\centering
\renewcommand{\arraystretch}{1.1} 
\setlength{\tabcolsep}{6pt}       

\newcolumntype{Y}{>{\centering\arraybackslash}X}

\begin{tabularx}{\textwidth}{|c|c|YYYY|YYYY|}
\hline
\multirow{2}{*}{\textit{Dataset}} 
 & \multirow{2}{*}{\textit{Model}} 
 & \multicolumn{4}{c|}{\textit{Character GLE} [\%] $\uparrow$}
 & \multicolumn{4}{c|}{\textit{Phoneme GLE} [\%] $\uparrow$} \\
\cline{3-10}
 & & Ours & \textsc{pwr} & \textsc{owa} & \textsc{lwa} & Ours & \textsc{pwr} & \textsc{owa} & \textsc{lwa} \\
\hline
\multirow{4}{*}{\makecell{\textsc{cv-en}}} & \textsc{whspr} & \textbf{78.8} & 77.0 & 65.8 & 58.9 & \textbf{74.2} & 73.1 & 60.6 & 54.2 \\
 & \textsc{phi4-m} & \textbf{78.6} & 76.8 & 66.0 & 59.8 & \textbf{74.2} & 73.2 & 61.2 & 55.2 \\
 & \textsc{pk-tdt} & \textbf{79.5} & 77.9 & 66.2 & 60.3 & \textbf{74.7} & 73.8 & 61.1 & 55.5 \\
 & \textsc{pk-ctc} & \textbf{77.0} & 75.2 & 65.9 & 59.7 & \textbf{72.6} & 71.5 & 61.3 & 55.4 \\
\hline
\multirow{4}{*}{\textsc{ted}} & \textsc{whspr} & \textbf{90.3} & 88.4 & 78.1 & 72.7 & \textbf{89.0} & 87.4 & 76.2 & 70.8 \\
 & \textsc{phi4-m} & \textbf{84.9} & 81.5 & 68.1 & 61.7 & \textbf{81.6} & 80.2 & 64.7 & 59.2 \\
 & \textsc{pk-tdt} & \textbf{87.6} & 84.9 & 74.0 & 68.7 & \textbf{86.0} & 83.8 & 71.8 & 66.8 \\
 & \textsc{pk-ctc} & \textbf{84.0} & 80.8 & 67.7 & 62.0 & \textbf{81.0} & 80.1 & 65.1 & 59.6 \\
\hline
\multirow{4}{*}{\textsc{pm57}} & \textsc{whspr} & \textbf{84.6} & 81.7 & 76.7 & 72.5 & \textbf{83.3} & 81.4 & 75.9 & 72.0 \\
 & \textsc{phi4-m} & \textbf{77.9} & 75.9 & 70.8 & 66.7 & \textbf{75.8} & 74.2 & 68.8 & 64.7 \\
 & \textsc{pk-tdt} & \textbf{79.4} & 77.2 & 71.7 & 67.3 & \textbf{77.8} & 76.2 & 70.3 & 66.0 \\
 & \textsc{pk-ctc} & \textbf{79.9} & 77.0 & 71.3 & 66.7 & \textbf{78.2} & 76.0 & 69.9 & 65.5 \\
\hline
\end{tabularx}
\caption{English-only evaluation across multiple models and datasets.\textsuperscript{$\dagger$}}
\vspace{-0.2em}
\label{tab:english}
\end{minipage}
\hfill
\begin{minipage}[t]{0.34\linewidth}
\centering
\renewcommand{\arraystretch}{1.1} 
\setlength{\tabcolsep}{6pt}       

\begin{tabularx}{\columnwidth}{|l|YYc|}
\hline
\multirow{2}{*}{\textit{Language}} 
 & \multicolumn{3}{c|}{\textit{Character GLE} [\%] $\uparrow$} \\
\cline{2-4} & Ours & \textsc{owa} & \textsc{lwa} \\
\hline
Portuguese  & \textbf{78.3} & 59.2 & 48.1 \\
Spanish     & \textbf{77.8} & 60.9 & 53.3 \\
Turkish     & \textbf{77.7} & 40.4 & 32.7 \\
German      & \textbf{76.9} & 47.0 & 40.7 \\
Polish      & \textbf{76.7} & 54.0 & 44.9 \\
Indonesian  & \textbf{76.5} & 56.5 & 49.5 \\
Swahili     & \textbf{73.9} & 45.3 & 34.4 \\
French      & \textbf{73.5} & 53.6 & 45.4 \\
\hline
\multicolumn{3}{|l|}{\textit{Ablations}: \textsc{cv-en} + \textsc{whspr}} & $\Delta$ \\
\hline
\multicolumn{3}{|l|}{Eq. \eqref{eq:cost} w/o substitution pen.} & -4.3 \\
\multicolumn{3}{|l|}{Eq. \eqref{eq:transition} w/ unit-cost} & -1.3 \\
\multicolumn{3}{|l|}{Search restricted to $G_b$} & -2.2 \\
\hline
\end{tabularx}
\caption{Multilingual eval. and ablations.\textsuperscript{$\dagger$}}
\vspace{-0.2em}
\label{tab:multilingual}
\end{minipage}
\end{table*}

\section{Evaluation procedure}
\label{sec:evaluation}

\subsection{Metric: Global-to-local edits (GLE)}
\label{ssec:metric}

We do not have access to human-annotated gold-standard alignments. Instead, we propose a distance measure between aligned text segments and apply reciprocal normalization using a theoretical lower bound on the full text. The distance measure is defined as
\begin{align}
d(r, h) = d_{\text{ID}}(r, h) + \text{abs}\left(|r| - |h|\right)\cdot \min(|r||h|, 1) \label{eq:metric} \, ,
\end{align}

\noindent where $d_{\text{ID}}(\cdot)$ denotes the edit distance restricted to insertions and deletions, and the second term adds the absolute length difference between $r$ and $h$ when both are non-empty.

Using only insertions and deletions prevents rewarding substitutions where the substrings contain no shared information, since an insert–delete pair receives the same score. The second term further prevents rewarding alignments for treating words as substitutions when they should be inserted.

To compute the GLE metric, we first sum $d_{\text{ID}}(\cdot)$ for all reference-hypothesis pairs in a dataset and divide it by the sum of $d(\cdot)$ over all individual alignments. We remove all unvoiced characters before evaluation, ensuring that the numerator and denominator are computed over identical strings.

\subsection{Baselines}
\label{ssec:baselines}

\textbullet\;\textit{Levenshtein-based word-level alignment} (\textsc{lwa}): The RapidFuzz toolkit \cite{rapidfuzz} provides one of the fastest implementations of Levenshtein distance computation, including extraction of alignments. The computed alignments represent an arbitrary optimal path through the Levenshtein backtrace graph.

\vspace{0.5em}

\noindent\textbullet\;\textit{Optimized word-level alignment} (\textsc{owa}): We implement a custom word-level alignment where transitions are scored according to \eqref{eq:metric}. In essence, this approximates a word-level oracle with reference to the proposed metric, and will help to highlight the limitations of simple one-to-one alignment.

\vspace{0.5em}

\noindent\textbullet\;\textit{Power alignment} (\textsc{pwr}): The Power aligner \cite{ruiz2015phonetically} realigns Levenshtein substitution spans after converting words to phonetic representations. The resulting alignments may span multiple words in the reference transcript, which can give a small advantage in relation to the proposed metric described above. While the Power aligner is the most relevant baseline, it only supports English, as it relies on manually curated linguistic resources for character-to-phoneme conversion. 

\vspace{-0.3em}
\subsection{Dataset and models}
\label{ssec:data_models}

We evaluate the alignment methods on three English datasets: Common Voice \cite{commonvoice2020}, TED-LIUM \cite{hernandez2018ted}, and PriMock57 \cite{korfiatis2022primock57}. Excluding the Power aligner, we evaluate on eight additional Latin-script languages from Common Voice selected based on speaker population size and linguistic diversity.

To ensure that the results generalize across model classes, we use four open-source speech recognition models: Whisper (v3, 1.6B) \cite{radford2023robust}, Phi-4-multimodal (5.6B) \cite{abouelenin2025phi}, Parakeet TDT (v2, 0.6B) \cite{rekesh2023fast}, and Parakeet CTC (0.6B) \cite{xu2023efficient}. For the non-English evaluation, we use only Whisper, as it is the only model that supports a sufficiently diverse set of languages.

\section{Results}
\label{sec:results}

As shown in Table~\ref{tab:english}, our method (beam size = 100) consistently outperforms all baselines at both the character and phoneme levels. Although the Power aligner is explicitly optimized for phonetic similarity, our approach achieves higher phoneme-level scores across every dataset and model. 

Table~\ref{tab:multilingual} (top) shows that our method also outperforms the baselines across all non-English languages. Relative improvements are generally larger for most non-English languages compared to English. 
We hypothesize that transcription quality for these languages makes alignment more challenging, which amplifies the benefits of our approach.

Table~\ref{tab:multilingual} (bottom) further highlights the benefit of key algorithmic choices. While each choice improves performance, the largest gain results from eliminating the one-to-one constraint of word-level alignment, discussed in section~\ref{ssec:word_level}, as seen by the difference between our approach and the \textsc{owa} baseline (-13.0, \textsc{cv-en} + \textsc{whspr}, Table \ref{tab:english}).

To verify these results, we ran a small post-hoc study in which two expert annotators indicated their preference between the two best-performing methods (ours and \textsc{pwr}) on 124 examples. The annotators showed a modest but significant preference for our method (p $\ll$ 0.01, binomial test). 

Finally, while our algorithm is slower than the optimized word-level methods ($\sim 10\times$ slower), it compares favorable to the Power aligner ($\sim 20\times$ faster).

\vspace{-0.5em}

\begingroup
\renewcommand\thefootnote{}
\footnotetext{\textsuperscript{$\dagger$}All results are significant (p $\ll$ 0.01, paired approx. permutation test).}%
\addtocounter{footnote}{-1}
\endgroup

\section{Conclusion}
\label{sec:conclusion}
\vspace{-0.6em}
We proposed a new alignment algorithm that significantly outperforms conventional methods across models, domains, and languages. The implementation is publicly released to support the community in developing and evaluating speech recognition systems.

\vfill\pagebreak

\bibliographystyle{IEEEbib}
\bibliography{refs}

\end{document}